\documentclass[journal]{IEEEtran}
\usepackage[pdftex]{graphicx}
\graphicspath{{../pdf/}{../jpeg/}}
\DeclareGraphicsExtensions{.pdf,.jpeg,.png}
\usepackage{mdwmath}
\usepackage{mdwtab}
\usepackage{eqparbox}
\usepackage{url}
\usepackage{booktabs,caption}
\usepackage{amsmath,amsfonts}
\usepackage{siunitx}
\usepackage{algorithmic}
\usepackage{algorithm}
\usepackage{array}
\usepackage{amsmath}
\usepackage[caption=false,font=normalsize,labelfont=sf,textfont=sf]{subfig}
\usepackage{textcomp}
\usepackage{stfloats}
\usepackage{url}
\usepackage{verbatim}
\usepackage{graphicx}
\usepackage{threeparttable}
\usepackage{cite}

\begin{document}

\title{Design of Whisker-Inspired Sensors for Multi-Directional Hydrodynamic Sensing}
\author{Tuo~Wang,~\IEEEmembership{Student Member,~IEEE,~ASME,} Teresa A.~Kent,~\IEEEmembership{Student Member,~IEEE,} Sarah~Bergbreiter,~\IEEEmembership{Member, IEEE, ~Fellow,~ASME}

\thanks{Tuo Wang and Sarah Bergbreiter are with the Department of Mechanical Engineering. Teresa Kent is with the Robotics Institute. All authors are with  Carnegie Mellon University, 5000 Forbes Ave, Pittsburgh, PA, 15213, USA. Corresponding author: Sarah Bergbreiter, sbergbre@andrew.cmu.edu, +1(412)268-3216}

}

\markboth{IEEE/ASME TRANSACTIONS ON MECHATRONICS}{}%

\maketitle

\begin{abstract}

Perceiving the flow of water around aquatic robots can provide useful information about vehicle velocity, currents, obstacles, and wakes. This research draws inspiration from the whiskers of harbor seals (\emph{Phoca vitulina}) to introduce a whisker-inspired water flow sensor. The sensor enables multi-directional flow velocity estimation and can be seamlessly integrated into aquatic robots. The whisker-inspired sensor operates using a mechano-magnetic transduction mechanism, which separates the whisker drag element from the electronic component. This configuration provides distinct advantages in terms of waterproofing and corrosion resistance. The sensor features a modular design that allows tuning of the whisker drag element's shape to optimize sensitivity and sensing range for diverse applications. An analytical model quantifies the sensor's capabilities, validated through experiments examining whisker parameters like morphology, cross-sectional area, aspect ratio, and immersion depth. The study also investigates the impact of structural designs on Vortex-Induced Vibrations (VIVs), an area of active research in both biological and robotic aquatic whiskers. Finally, the sensor's efficacy is demonstrated on a commercially available, remotely controlled boat, highlighting its ability to estimate flow velocity. The presented sensor holds immense potential for improving aquatic robots' navigation and perception capabilities in a wide range of applications.

\end{abstract}

\begin{IEEEkeywords}

whisker-inspired designs, flow sensing, underwater robots, flow sensing, bio-inspired robotics

\end{IEEEkeywords}

\section{Introduction}


The field of aquatic robotics has made significant strides in recent years, resulting in the development of various sophisticated machines such as remotely controlled ships \cite{neal2012hardware,kasda2021low,jo2019low}, underwater autonomous vehicles \cite{8706541,yoerger2021hybrid,wang2020development}, and soft underwater robots \cite{katzschmann2018exploration,teoh2018rotary,patterson2020untethered,aubin2019electrolytic}. These robots have showcased impressive capabilities across a diverse range of applications, including environmental monitoring \cite{tuhtan2020underwater,leonard2010coordinated}, search and rescue missions \cite{4749579,9987047}, and ocean exploration \cite{picardi2020bioinspired,8706541,6588608}. Despite their success, the majority of these systems have no way to measure fluid flow around the vehicle -- a measurement that can provide useful information regarding the vehicle's velocity, currents, nearby obstacles, and wakes \cite{dehnhardt_hydrodynamic_2001, eberhardt_development_2016, jiang_underwater_2022}. For small vehicles in particular, it is especially challenging to sense fluid flow around the vehicle.

Looking to nature, we find many aquatic animals, such as harbor seals and seal lions, have evolved biological flow sensors known as whiskers (vibrissae) to actively measure their surrounding flow field for foraging and navigation \cite{Whisker,adachi2022whiskers}. A biological whisker sensor contains a high-aspect-ratio whisker drag element that displaces when the animal moves. The displacement is then transmitted through densely located sensory cells embedded in the whisker root and is finally processed as surrounding flow information. Compared with human-engineered navigation systems like cameras and GPS, whisker sensing does not require ideal lighting conditions and is not subject to attenuation issues \cite{lee2012vision,tan2011survey,kim_soft_2023}. Another benefit of whisker sensing is that all neurological sensing happens at the whisker's base; since the whisker drag element is a specialized hair, damage to the whisker does not affect the sensors. This principle of separation enhances the durability and reliability of the sensors in engineering applications. As a result, the sensor remains functional even when the drag element is damaged, enhancing its durability and reliability in engineering applications.



The idea of using a whisker-inspired sensor to measure water flow has already been realized using many different sensing modalities, such as piezoresistive \cite{liu2022artificial, zheng20223d, piezoresistivesensor2}, capacitive \cite{eberhardt2016development}, magnetostrictive \cite{na2018magnetostrictive, 6971622}, and triboelectric \cite{xu2021triboelectric}. Common techniques like 3D printing \cite{gul2018fully}, molding {\cite{wang2022underwater, piezoresistivesensor2}, laminating \cite{xu2022bio}, and microelectromechanical systems (MEMS) manufacturing \cite{6765747} help create these sensors to fit different robot sizes. These whisker-inspired sensors have been applied in stationary setups to measure oscillatory flow \cite{liu2022artificial} and on mobile robots to detect wakes \cite{beem_calibration_2013}. Despite success in specific applications, challenges remain to whisker-inspired sensors' widespread application.



One challenge associated with aquatic whisker-inspired sensors is waterproofing. In earlier sensor designs, the whisker drag element (as shown in Fig. \ref{sensingMechanism}) was typically in direct contact with an electronic component, like a strain gauge \cite{zheng20223d, liu2022artificial} or capacitor \cite{capacitivewhiskersensor1, capacitivewhiskersensor2}. This meant that both the electronic part and the joint connecting it to the whisker drag element had to be waterproofed. Previous methods addressed this problem through mechanical sealing \cite{6404978, 6697220}, or coating with waterproof materials \cite{li2017superhydrophobic, qualtieri2012parylene}. However, during flow sensing the waterproofing layer can crack and delaminate due to vibration and rotation of the whisker drag element.

\begin{figure}[t]
   \begin{center}
  \includegraphics[width=3.5in]{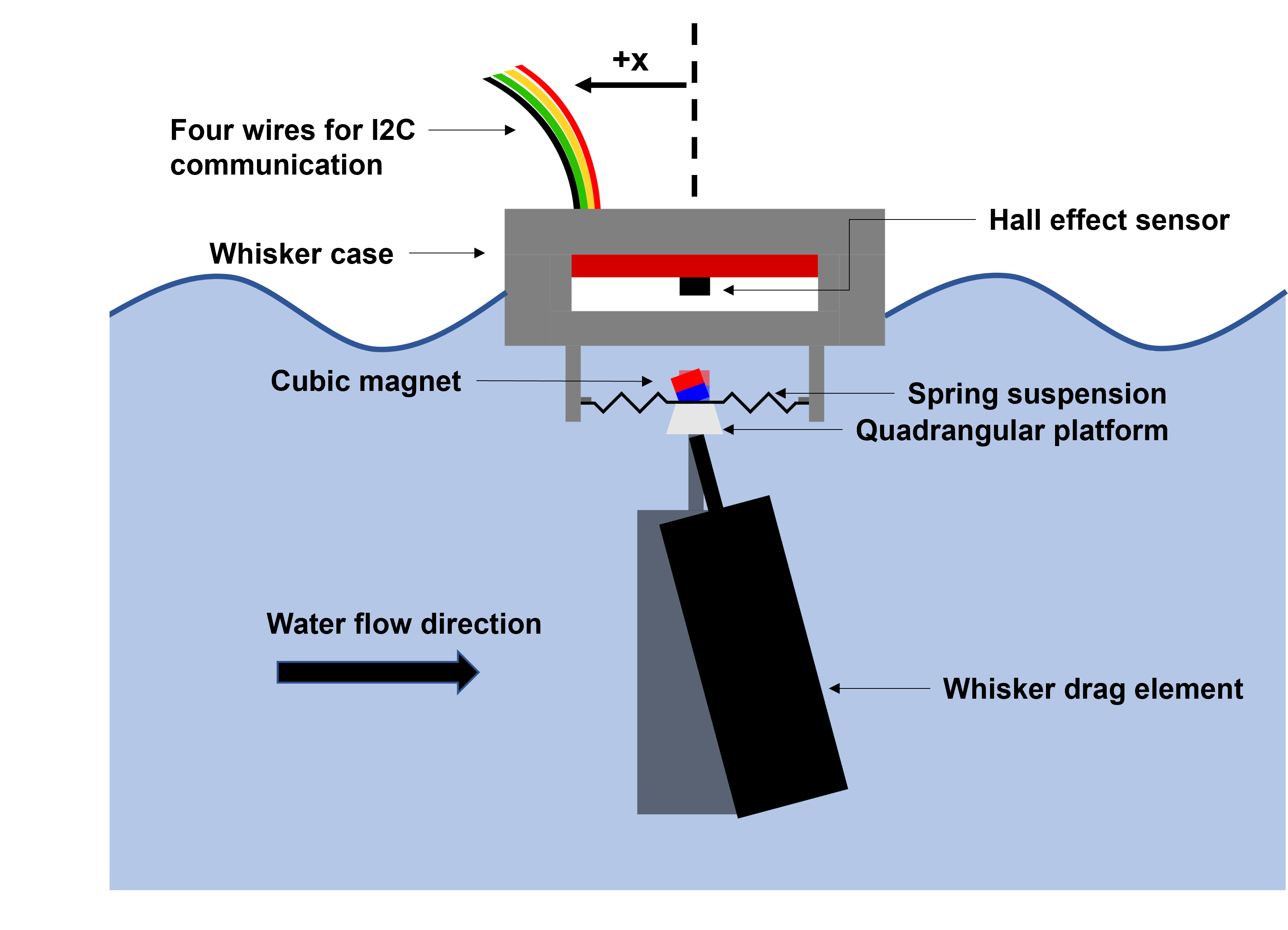}
  \caption{The sensing mechanism of the whisker-inspired sensor. As the holder rests on the water's surface and moves towards the left in the image, the whisker drag element rotates to the right due to induced flow.  This rotation to the right results in a counterclockwise rotation of the connected system, thereby altering the magnet's position in relation to the Hall effect sensor. The magnetic field response is then captured by the Hall effect sensor. The positive x-axis indicates a positive sensor reading for the magnetic field response (+x).}
  \label{sensingMechanism}
  \end{center}
\end{figure}

Another challenge lies in distinguishing mixed hydrodynamic signals in flow sensing. Whiskers not only detect relative water flow velocity, but also experience Vortex-Induced Vibrations (VIVs) caused by vortices shedding downstream of the whisker drag element, resulting in periodic fluid forces on the whisker. The frequency and strength of VIVs are closely related to the fluid flow velocity \cite{VIV1,VIV2}. While earlier research aimed to understand the seal whisker's shape to minimize VIVs for noise reduction, both biology and engineering studies revealed that VIVs can actually aid animals or robots in locating objects without vision or sonar systems \cite{whiskersurpressVIV1, whiskersurpressVIV2, sensorreview1, zheng2023wavy}. As a result, developing flow sensors that can differentiate and simultaneously capture water flow and VIV signals could lead to a considerably more effective sensor.




To overcome these challenges, this work presents the design of waterproofed, magnetically-transduced, whisker-inspired sensors for multi-directional hydrodynamic sensing. The magnetic transduction mechanism requires no physical connection between the sensing electronics and the whisker drag element (Fig. \ref{sensingMechanism}), which simplifies the waterproofing of the sensor. By water-cutting carbon fiber sheets, the whisker drag element can be efficiently manufactured and characterized, making this rapid prototyping technique a powerful tool for creating various whisker profiles. In this work, we test three different whisker drag element profiles on our sensor: rod, plate and cross shapes. Through rigorous characterization and modeling, we provide insights into the relationships between whisker shapes, vortex-induced vibration (VIV) responses, and sensing capabilities. This information provides a solid foundation for future sensor designs and applications. Additionally, we demonstrate our whisker-inspired sensor's performance for velocity estimation on a small, off-the-shelf remote-controlled boat, highlighting its potential for advancing aquatic robotics and sensing technologies.



\begin{figure}[t]
   \begin{center}
  \includegraphics[width=2.3in]{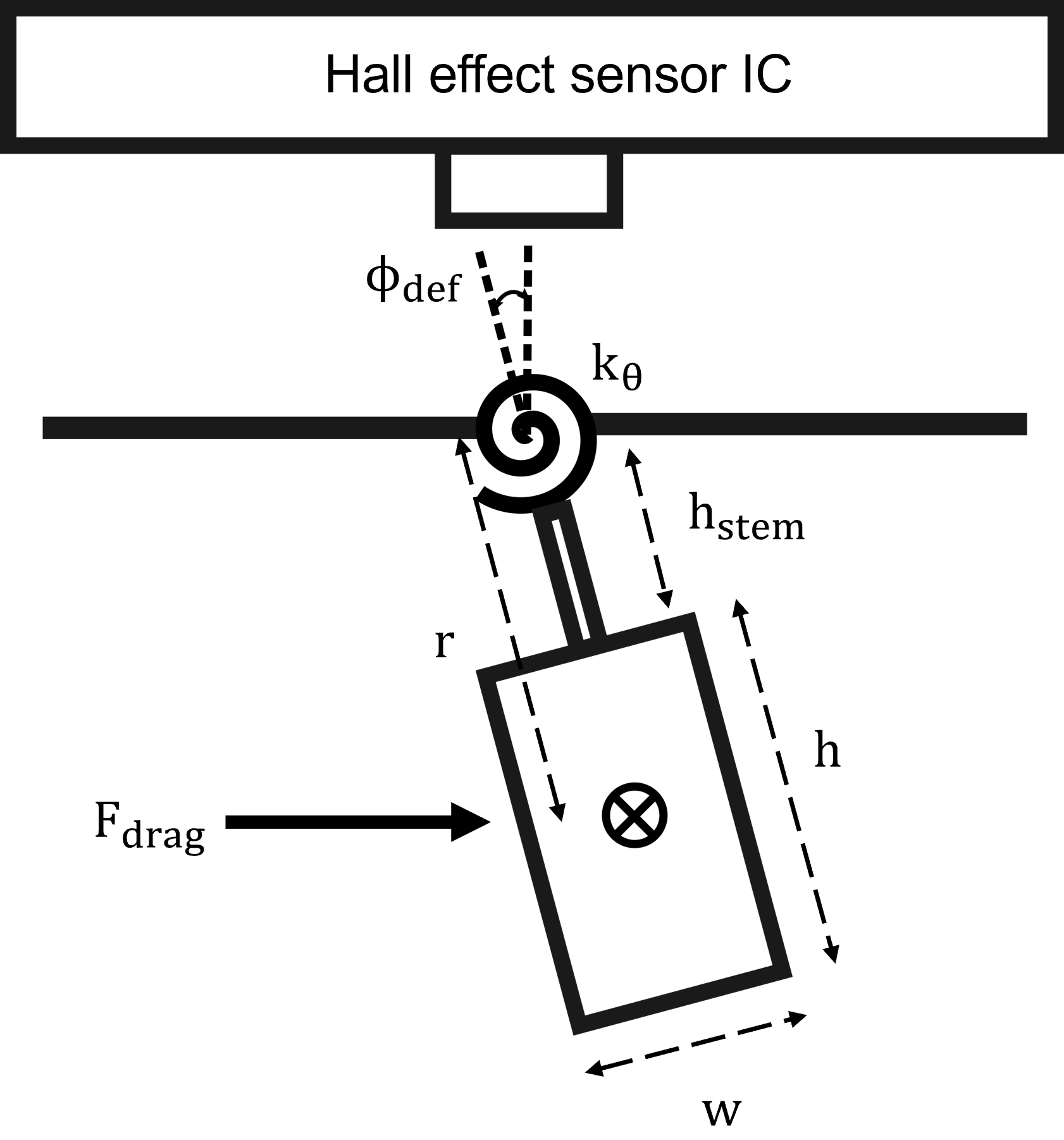}
  \caption{A schematic illustrating the torsional spring model for the whisker-inspired sensor. $F_{drag}$ is the dragging force induced by the relative water flow. $k_{\theta}$ is the torsional spring constant of the spring suspension. $\phi_{def}$ is the polar deflection angle of the center plate. $r$ is the moment arm of the dragging force from the lumped force center to the center plate. $h_{stem}$ represents the distance from the upper side of the whisker drag element to the spring suspension, and $h$ and $w$ are the height and width of the whisker drag element, respectively.}\label{modelSchematic}
  \end{center}
\end{figure}

\section{Whisker-inspired Sensor System Design}

Fig. \ref{sensingMechanism} depicts the sensing schematic and design of the whisker-inspired sensor, which is based on the design in \cite{MRLwhisker01}. The sensor system is comprised of a whisker drag element that is constrained by a spring suspension. A \SI{2}{\cubic\milli\metre} permanent magnet is located on the opposite side of the spring from the whisker drag element. The Hall effect sensor and its electronics are located separately from the whisker drag element and water. They are positioned in a way that ensures symmetry in the magnetic field response, regardless of the direction of rotation. This is achieved by placing them in line with the center of the magnet.

Separating the whisker drag element from the Hall effect sensor allows the whisker drag element to make contact with the water, while protecting the transduction part in a water-tight encasing. As water flows over the whisker drag element, the drag force induced by the flow deflects the whisker, causing the magnet to move in the opposite direction of the flow. Consequently, the motion of the magnet causes a variation in the magnetic field, which is detected by the Hall effect sensor. In addition to its waterproofing benefits, the sensor design comprises components with previously established models as part of a tactile sensor \cite{MRLwhisker01}, and these models can be used to predict the expected sensor response to water flow.


\subsection {Sensor Modeling}
\begin{figure*}[t]
   \begin{center}
  \includegraphics[width=6in]{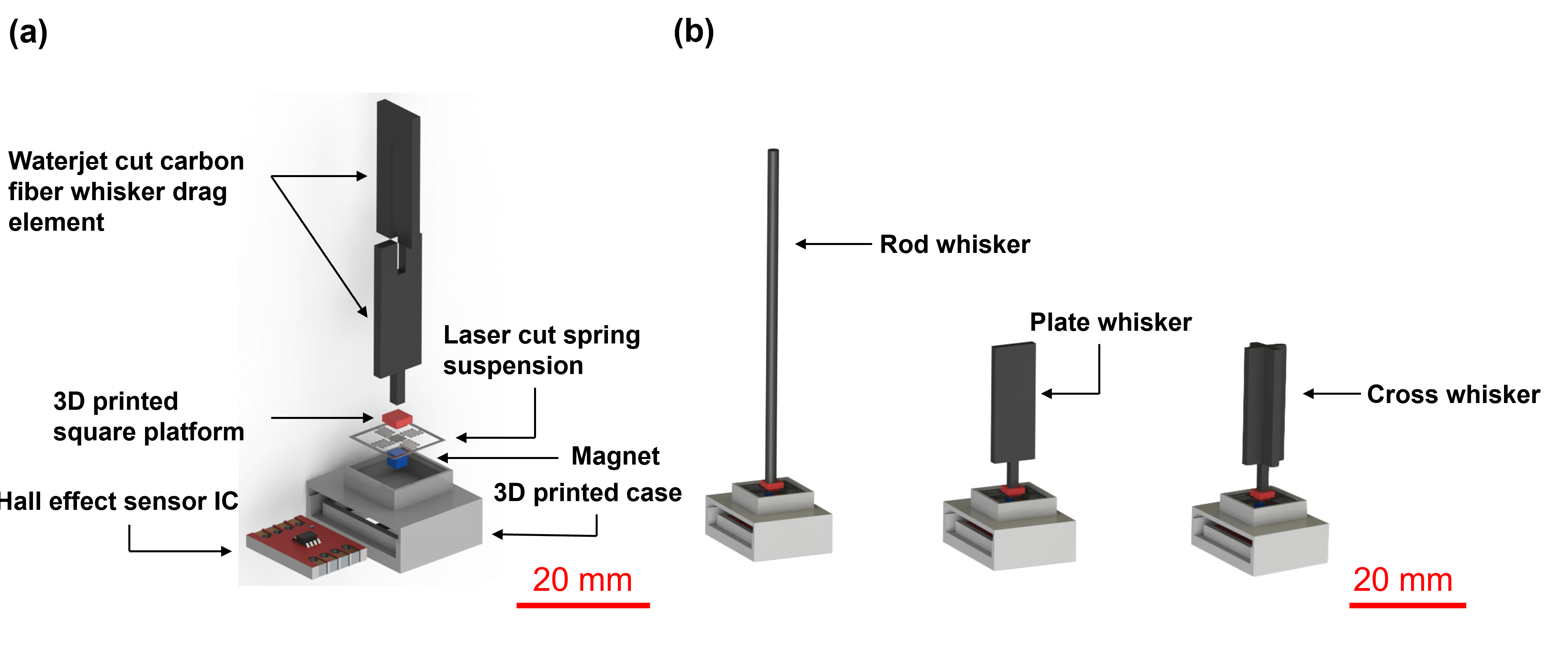}
  \caption{The fabrication overview of the whisker-inspired sensor. (a) The components and assembly of the sensor. The red square platform and blue magnetic cube are shown in false color to make these components distinguishable from the graph. (b) Three different whisker morphologies, from left to right: rod, plate, and cross whisker.}\label{fabrication}
  \end{center}
\end{figure*}

The model of the sensor focuses on the relationship between the flow velocity ($v$) experienced by a whisker drag element and the magnetic field response ($B_x$, $B_y$, $B_z$) sensed by the Hall effect sensor. In this paper, we assume that water flowing at a constant velocity produces a constant drag force on a whisker drag element. Given that the whisker drag element is fixed to a spring at one end, the whisker will rotate until it has reached a quasi-static angle where the moment induced by the drag force and the opposite moment induced by the spring rotation are equal (Fig. \ref{modelSchematic}). Here the angle of the balanced moments is named the quasi-static polar deflection angle ($\phi_{def}$, Fig. \ref{modelSchematic}). $\phi_{def}$ is also the angle of the magnet relative to the z-axis and thus determines the change in the magnetic field sensed by the Hall effect sensor. 

The whisker-inspired sensor model combines models for the expected drag force and moment (Section \ref{dragModel}), the expected spring response to applied moments (Section \ref{springDesign}) and changes in the magnetic field caused by a rotating magnet (Section \ref{magnetDesign}) to model a unique relationship between the magnetic field and flow velocity. The model's success is evaluated by its ability to predict a manufactured sensor design's maximum flow velocity, flow velocity at mid-range, and sensitivity measured in LSB/\SI{}{\milli\metre\per\second}.

\subsubsection{Drag Moment Caused by Flow}
\label{dragModel}
The drag force on a whisker ($F_{drag}$) moving through a fluid of density $\rho$ at a velocity $v$ is modeled by the drag equation (Eqn. \ref{dragEquation}). In Eqn. \ref{dragEquation} the shape of the whisker drag element determines the drag coefficient ($C_d$), and the drag element's height ($h$), width ($w$), and thickness ($t$) are factors in the cross-sectional area ($A$). $\rho_{water}$ is the density of the water that the whisker is moving through.

\begin{equation}
F_{drag} = \frac{1}{2} C_d\rho_{water}v^{2}A
\label{dragEquation}
\end{equation}

It is important to note that as the direction of flow relative to the sensor axis ($\theta_{xy}$) changes, $C_d$ and $A$ can also change for the three drag element designs in Fig. \ref{fabrication}. For example if $A_{0}$ is the projection area at $\theta_{xy}$ = \SI{0}{\degree}, the projected area at angle $\theta_{xy}$ ($A_{\theta}$) can be calculated using Eqn. \ref{referenceArea}. 

\begin{gather}
    A_{\theta} = A_{x0} cos(\theta)+A_{y0} sin(\theta)
    \label{referenceArea}
\end{gather}

The moment ($M_{drag}$, Eqn. \ref{TorsionSpring}) exerted at the whisker-spring connection combines the force found in the drag equation (Eqn. \ref{dragEquation}) and the magnitude of the moment arm ($r$, Fig.\ref{modelSchematic}) calculated in Eqn. \ref{MomentArm}. In this model, we assume uniform flow perpendicular to the sensor and a uniform cross-sectional area of the whisker, but we do not assume the whisker is fully immersed. While full immersion is required for underwater vehicles, surface aquatic vehicles can potentially tune this parameter when adding the sensor to the vehicle. We use the variable $D_{im}$ (illustrated in Fig. \ref{characterizationSetup}(d)) to indicate the distance beneath the water surface that the drag element is immersed. 

\begin{equation}
r = h_{stem}+h - \frac{1}{2}D_{im}
\label{MomentArm}
\end{equation}
\begin{equation}
M_{drag} = F_{drag}r
\label{TorsionSpring}
\end{equation}

\subsubsection{Spring Design}

\label{springDesign}
The model of the spring suspension design in Fig. \ref{fabrication} is based on previous work by Chou et al. \cite{springmodel} that uses Castigliano's beam theory to model four connected serpentine springs. This spring model was previously verified in tactile sensing and airflow estimation tasks in \cite{MRLwhisker01,MRLwhisker02}. For the rigid whiskers in this work, rotation of the whisker drag element exerts a rotation of equal magnitude on a center plate connecting the four springs. This rotation is described by the magnitude of rotation ($\phi_{def}$,  Fig. \ref{modelSchematic}) and direction of rotation ($\theta_{xy}$, Fig. \ref{characterizationSetup}(b)). Previous work \cite{MRLwhisker01} has shown that when the spring parameters are known, moments about the x and y axis ($M_x$ and $M_y$) correlate to a unique pair of values for $\theta_{xy}$ and $\phi_{def}$ based on inverse elements of the compliance matrix (Eqns. \ref{M_x},\ref{M_y}).

\begin{gather}
    M_x \propto sin(\phi_{def})cos(\theta_{xy})
    \label{M_x}\\
    M_y\propto sin(\phi_{def})sin(\theta_{xy})
    \label{M_y}
\end{gather}

The final spring parameters used are similar to those found in \cite{MRLwhisker01, MRLwhisker02} and these parameters are repeated in Table \ref{spring} for completeness. The code and detailed mathematical derivations for this work are available on GitHub \cite{Kent2022}. While this spring model is not novel on its own, we have added $\pm$\SI{20}{\degree} limits to the maximum $\phi_{def}$ in the model to represent physical mechanical stops that prevent the spring suspension from plastically deforming. The spring design was considered fixed in the model used in this paper, but could ultimately be varied to adjust sensing range in future work.

\subsubsection{Hall Effect Sensing}
\label{magnetDesign}

The calculations for converting $\phi_{def}$ into the magnetic field response were previously published in \cite{MRLwhisker01}. Here we provide a brief description for completeness and describe our incorporation of the Hall effect sensor's limitations into the sensor design model. The magnet rotates about the spring suspension's center by the same $\phi_{def}$ and $\theta_{xy}$ as the whisker drag element does (Fig. \ref{sensingMechanism}). Derby and Olbert's equations \cite{magneticmodel} describe the expected changes in magnetic flux at a point when a magnet rotates and translates relative to that point. The Hall Effect sensor used in this work provides magnetic field measurements in 3 axes, and we specifically model the change in magnetic field around the x and y axes -- $\Delta B_x$ and $\Delta B_y$.

In the current work, we have added the resolution and sensitivity of the Hall effect sensor to our model. The Hall Effect sensor used in this work has a published sensitivity of \SI{5}{} Least Significant Bits (LSB) per \SI{}{\milli\tesla} and a maximum magnetic field range of \SI{230}{\milli\tesla}. In order for our sensor design to detect a rotation applied to the magnet, the rotation must result in a change of magnetic field of at least \SI{0.2}{\milli\tesla}. The model also indicates that the sensitivity of the sensor to flow velocity can be increased by either reducing the distance between the magnet and the sensor, or by increasing the magnetization of the whisker magnet. 


\subsubsection{Design Modeling}
\begin{figure*}[t]
   \begin{center}
      \includegraphics[width=6in]{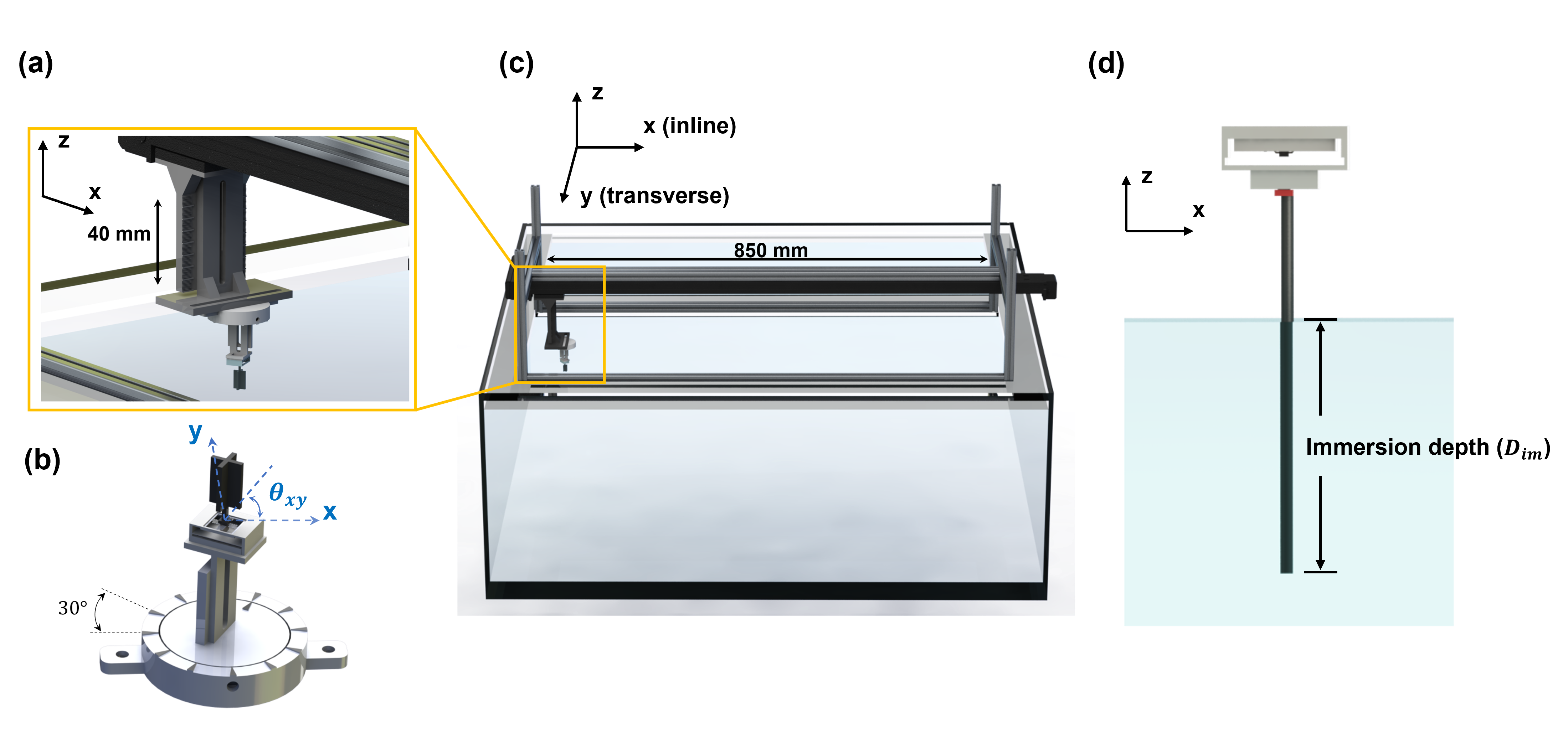}
   \vfill
  \caption{Characterization Setup with the water tank. (a) A 3D printed holder that adjusts the depth of the sensor in water. (b) The 3D printed sensor mount that adjusts orientations of the sensor. (c) The setup of the linear stage and water tank. (d) Diagrammatic image showing the immersion depth vs height for a rod whisker.}
  \label{characterizationSetup}
  \end{center}
\end{figure*}

 Decisions regarding the design of the whisker drag element, the level of immersion of the whisker drag element in water, the parameters of the spring, the magnet, and Hall effect sensor all play a role in determining the sensor response.  In this paper we focus our study on how changes in the whisker drag element design and immersion in the water affect the whisker-inspired sensor's sensitivity to flow velocity. By combining Eqn. \ref{TorsionSpring} with the spring model, we can analyze the effect of different whisker drag element designs on the anticipated rotation angles $\phi_{def}$. 
 
 Three cross-sectional shapes of the whisker drag element illustrated in Fig. \ref{fabrication}(b) were considered in the model: a rod, plate and cross. The three whisker shapes were modeled with drag coefficients of 1.1, 1.32, 1.32 for the rod, plate and cross, respectively (fit to experimental results). It is worth noting that we only modeled the cross and the plate in one $\theta_{xy}$ direction, and other $\theta_{xy}$ directions will have different $C_d$ and $A$ values for these shapes. The model results of the sensors were compared to the experimental results for each of the three designs with varying the immersion depth and orientation.

 
 The model was further compared to experimental data for the rod whisker across different areas and immersion depths. Immersion depth provides an opportunity to adjust the sensor's sensitivity and range during application in contrast to the drag element shape that is fixed during fabrication. We used the model to better understand the sensitivity of the whisker sensor to both the 'hard-coded' design parameter of rod diameter and the adjustable design parameter of immersion depth. We evaluated the sensor's sensitivity (measured in LSB/\si{\milli\metre\per\second} and evaluated at the sensing range mid-point due to nonlinearity), mid-point flow velocity, and maximum flow velocity using the model and compared the modeled design space to experimental results. 
 

\subsection{Fabrication}

We developed an efficient assembly process to create versatile sensors featuring three distinct drag element shapes of various sizes, enabling us to effectively evaluate their performance. Fig. \ref{fabrication} shows the fabrication and assembly process of the sensor. The plate and cross whiskers were fabricated by waterjet (ProtoMax Abrasive Waterjet, OMAX) using a \SI{1.5}{\milli\metre} thick carbon fiber sheet (StarimCarbon). The cross whiskers were then mechanically assembled into shape. The rod whiskers were made by cutting carbon fiber rods (Awclub) of \SI{1}{\milli\metre}, \SI{2}{\milli\metre}, and \SI{3}{\milli\metre} diameters. Unlike biological whiskers which are compliant and tapered, the fabricated whiskers are rigid and have constant cross-sectional areas along their height. To improve the adhesion and alignment of the whisker on the center plate, we 3D printed a \SI{1.5}{\milli\metre} high square platform (Object 30, Stratasys). This platform was then glued to the bottom of the whisker assembly before it was affixed to the spring suspension. 

The spring suspension was made by laser cutting (Photolaser U4, LPKF) a \SI{100}{\micro\metre} thick stainless steel sheet (Shop-Aid Inc.). Laser settings for power (\SI{2}{\watt}) and repetitions (\SI{80}) were set to increase the cutting resolution. The spring suspension was then soaked in \SI{90} percent isopropanol solution for \SI{10} minutes to reduce the cutting residuals on its surface. A \SI{2}{\cubic\milli\metre} cube magnet (C0020, Supermagnetman) was glued to the back side of the spring. Once assembled, the spring suspension and whisker drag element were press-fit and then glued into the 3D-printed case.

The Hall effect sensor (TLE493-W2B6 A0, Infineon Technologies) was also press-fit into the 3D-printed case (Object 30, Stratasys). Four wires were soldered to the board through a hole at the bottom of the whisker case for I2C communication. Finally, silicone adhesive (Sil-poxyTM, Smooth-On) was applied to waterproof the sensor. Table~\ref{whiskerSizes} lists the dimensions of all nine whisker shapes used in modeling and characterization.


\begin{table}[h]
    \caption{Spring Parameters}
    \label{spring}
    \centering
\scalebox{1}{
    \begin{tabular}{c|c}
    \hline
    Parameters & Value \\
    \hline
    \hline
    Material &  Stainless Steel \\
    Thickness (t) & \SI{100}{\micro\meter} \\
    Arm Width (w) & \SI{0.25}{\milli\meter} * \\
    Length (l) & \SI{1}{\milli\meter} \\
    Pitch (p) & \SI{0.41}{\milli\meter} \\
    Turn (n) & \num{6} \\
    \end{tabular}}
    \begin{tablenotes}
        \item *: Due to the LPKF laser cutting process, the width of the upper and lower sides of the spring is slightly different.  A pictorial representation of these spring parameters can be found in \cite{MRLwhisker01}.

    \end{tablenotes}
\end{table}

\begin{table}[h]
    \caption{Dimensional Parameters of Whisker Drag Elements}
    \label{whiskerSizes}
    \centering
\scalebox{1}{
    \begin{tabular}{c|c|c}
    \hline
    Whisker & Height (mm) & Width / Diameter (mm) \\
    \hline
    \hline
    Cross1 & 30 & 5\\
    Cross2 & 20 & 7.5\\
    Cross3 & 15 & 10\\
    Plate1 & 30 & 5\\
    Plate2 & 20 & 7.5\\
    Plate3 & 15 & 10\\
    Rod1 & 60 & 1\\
    Rod2 & 60 & 2\\
    Rod3 & 60 & 3\\
    \end{tabular}}
\end{table}

\begin{figure*}[h]
   \begin{center}
  \includegraphics[width=6in]{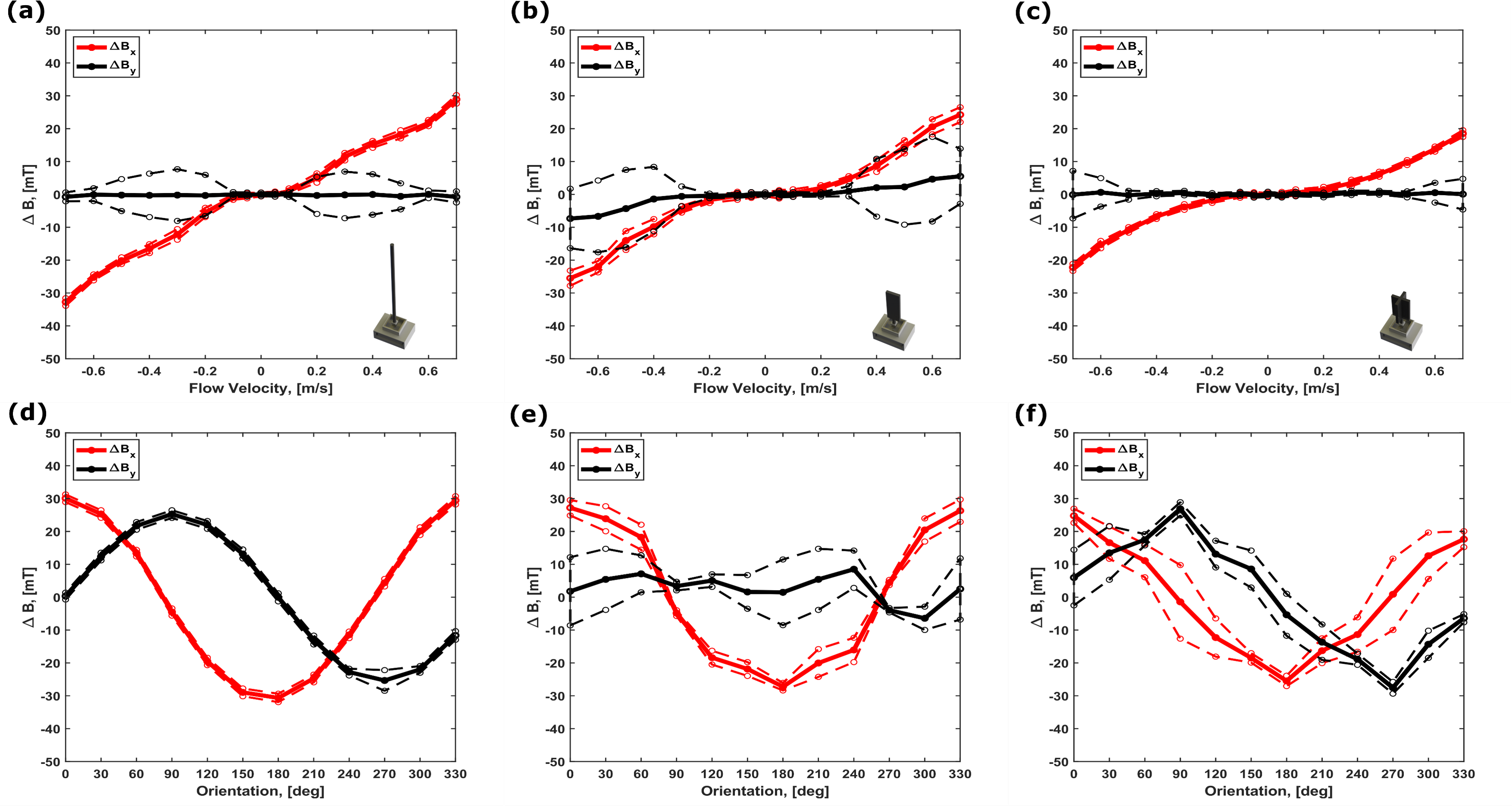}
  \caption{The characterization of the sensors' magnetic field response is analyzed across three different whisker drag element shapes. Arranged by column, the shapes are (a)(d) \SI{3}{\milli\metre} diameter rod whisker, (b)(e) \SI{30}{\milli\metre}$\times$\SI{5}{\milli\metre} plate whisker, and (c)(f) \SI{30}{\milli\metre}$\times$\SI{5}{\milli\metre} cross whisker. In (a-c), the magnetic field response is divided into two components: the red x-direction and the black y-direction (Fig.\ref{characterizationSetup} (c)). The dotted belt signifies one standard deviation of the magnetic field response gathered during experiments. In (d-f), the magnetic field response is examined as the orientation angle between the in-line flow and the sensors' x-axis changes at a fixed flow velocity of \SI{0.7}{\metre\per\second}.}
    \label{characterizationResults}


  \end{center}
\end{figure*}

\subsection{Characterization Setup}

\label{characterizationSection}


To assess the impact of drag element geometry on sensitivity and detection range of whisker-inspired sensors, several experiments were conducted. These tests involved subjecting the sensors to different flow velocities and orientations while immersed in water. To accomplish this, a linear stage (X-BLQ-E1045, Zaber Technologies Inc.) was mounted on a water tank (\SI{1200}{\milli\metre} × \SI{1200}{\milli\metre} × \SI{440}{\milli\metre}) and supported by customized aluminum T-slot beams (25 Series, 80/20 Inc.). The linear stage was programmed to follow prescribed velocities with a fixed acceleration profile.

A 3D-printed sensor mount (Object 30, Stratasys) and two holders (Raise3D Pro2, Raise3D) were utilized to secure the sensor to the linear stage, enabling adjustment of the sensor's orientation and immersion depth within the water. The sensor was then mounted on the linear stage and dragged through static water, generating a relative water flow. The experimental setup is depicted in Fig. \ref{characterizationSetup}.

Data from the whisker-inspired sensor's response to motion was acquired by an Arduino Mega 2560 R3, using I2C communication at a sampling frequency of 100 Hz. For each trial, the sensor was moved along a linear stage within the water tank for a total travel distance of 850 mm, with a consistent acceleration and deceleration of \SI{2}{\metre\per\second\squared} at the beginning and end of the run. This process ensured a minimum of 70 data points were collected at the prescribed velocity per trial. Mean and standard deviation values were calculated from the magnetic field response in each trial and used to analyze the relationship between flow velocity and vortex-induced vibration (VIV) signals.

\subsubsection{Varying whisker morphologies}

Each of the nine whiskers (described in Table~\ref{whiskerSizes}) was characterized with eight flow velocities, [\SI{0.05}, \SI{0.1}, \SI{0.2}, \SI{0.3}, \SI{0.4}, \SI{0.5}, \SI{0.6}, \SI{0.7}] \SI{}{\metre\per\second}, and twelve $\theta_{xy}$ orientations, [\SI{0}{\degree}, \SI{30}{\degree}, \SI{60}{\degree}, \SI{90}{\degree}, \SI{120}{\degree}, \SI{150}{\degree}, \SI{180}{\degree}, \SI{210}{\degree}, \SI{240}{\degree}, \SI{270}{\degree}, \SI{300}{\degree}, \SI{330}{\degree}]. For each combination of velocity and orientation, four trials were conducted.


\subsubsection{Varying immersion depths for rod whiskers}


Fig. \ref{characterizationSetup}(d) provides a more detailed schematic illustrating how the response of rod whiskers with diameters of \SI{1}{\milli\metre}, \SI{2}{\milli\metre}, and \SI{3}{\milli\metre} were characterized at different immersion depths using this same characterization setup. The rod whisker was mounted perpendicular to the water surface, and a portion of its height was immersed in water. The total height of all of the rod whiskers was \SI{60}{\milli\metre}, and the immersion depths ranged from \SI{10}{\milli\metre} to \SI{40}{\milli\metre} in \SI{10}{\milli\metre} increments.


\begin{figure*}[t]
   \begin{center}
  \includegraphics[width=6in]{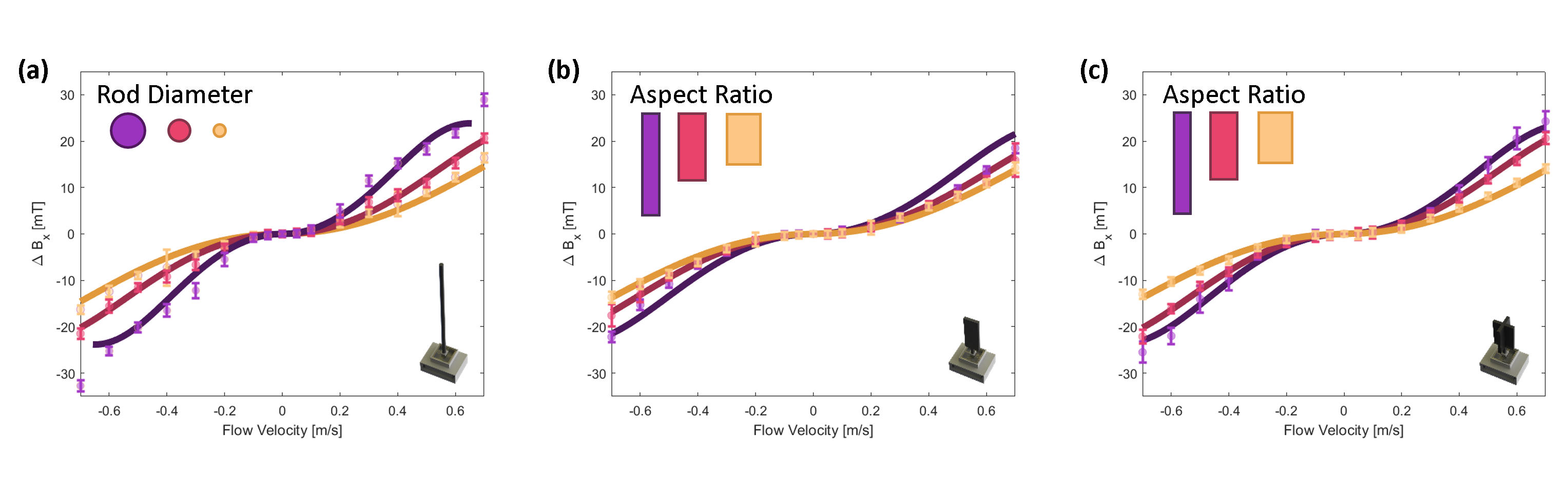}
  \caption{Comparison of the modeled values of $B_x$ with the experimental results for the: a) rod whisker, b) plate whisker, c) cross whisker. The experimental data is from testing each of the whiskers in Table \ref{whiskerSizes}. The colors represent three sensors with different heights and widths. In a) the different colors represent the different whisker diameters with purple, magenta and yellow represeting a diameter of 3, 2, and 1 \SI{}{\milli\meter} respectively. In (b) and (c) the area between each whisker is consistent but the moment arm (r) changes: 15, 10, 7.5 \SI{}{mm} for purple, magenta and yellow.} \label{Modeling}
  \end{center}
\end{figure*}

\begin{figure*}[t]
   \begin{center}
  \includegraphics[width=6in]{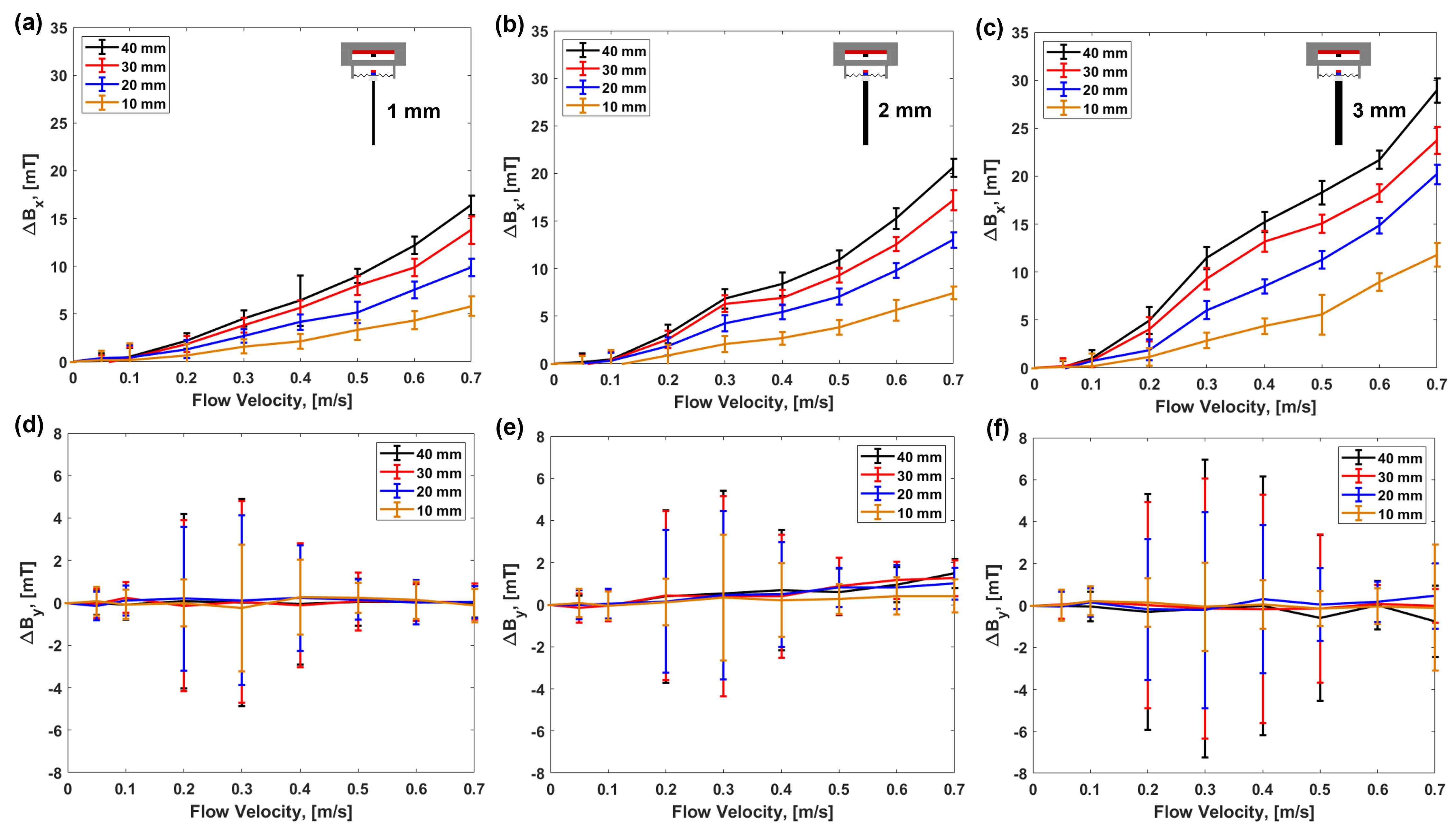}
  \caption{Characterization of the magnetic field response for \SI{1}{\milli\metre} diameter, \SI{2}{\milli\metre} diameter, and \SI{3}{\milli\metre} diameter whiskers with four different immersion depths (Fig.\ref{characterizationSetup} (d)) in water.  (a) (b) (c) represent the magnetic field response along the in-line direction for \SI{1}{\milli\metre} diameter, \SI{2}{\milli\metre} diameter, and \SI{3}{\milli\metre} diameter whisker-inspired sensors, respectively. (d) (e) (f) represent the magnetic field response in the transverse direction for \SI{1}{\milli\metre} diameter, \SI{2}{\milli\metre} diameter, and \SI{3}{\milli\metre} diameter whisker-inspired sensors, respectively. Error bars represent one standard deviation from four trials.
  } \label{depthcharacterization}
  \end{center}
\end{figure*}
\begin{figure*}[t]
   \begin{center}
  \includegraphics[width=6in]{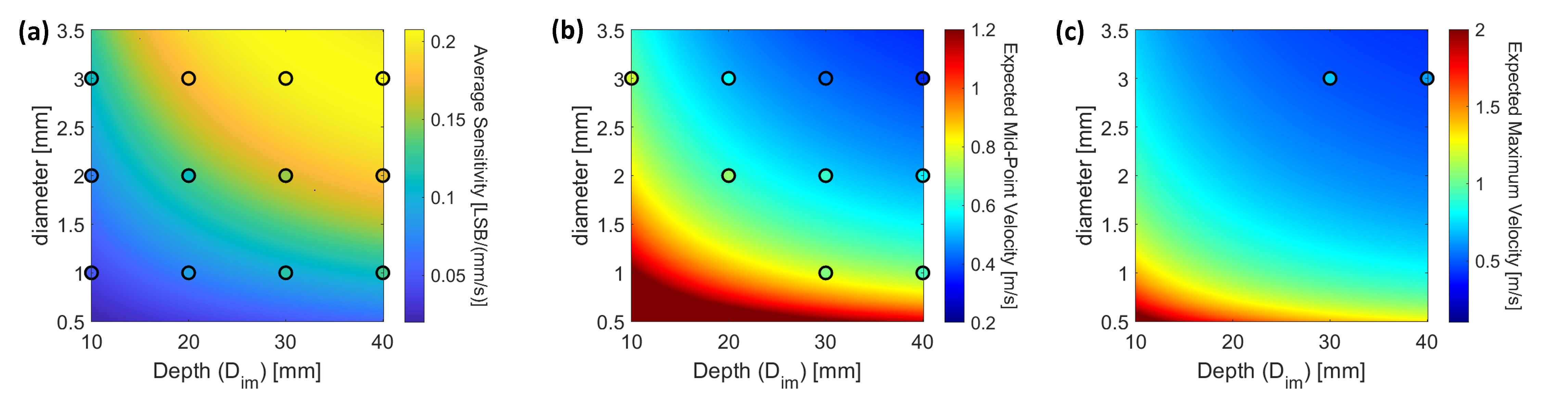}
  \caption{Modeling and experimental work showing the effect of the depth of immersion and radius of the rod whisker on sensor specifications. The importance of each specification is described in Section II.A.4. The heat map shows the model results. Results from the experimental tests are shown in circles, with the colors representing the experimental values. Three sensor specifications are considered: a) The Sensitivity: The average sensitivity of the sensor to flow between the speeds of \SI{0.2}{\metre\per\second} and \SI{0.7}{\metre\per\second} flow. b) Sensor Mid-Point Velocity: The modeled and measured velocity at \SI{12.5}{\degree} $\phi_{def}$ rotation c) Maximum Detectable Velocity: The maximum velocity value the sensor can achieve without damage to the sensor.}
  \label{Design Decisions}
  \end{center}
\end{figure*}

\subsection{Sensor calibration for surface vehicle demonstration}

A \SI{20}{\milli\metre} $\times$ \SI{7.5}{\milli\metre} cross sensor was chosen for use on a remotely-controlled (RC) boat based on this sensor's sensitivity and sensing range. The sensor was calibrated for velocity and orientation so that it could ultimately be used for velocity estimation on-board the boat. In this calibration, the flow velocity ($v$) was estimated from the magnetic field responses in the x and y directions, $\Delta B_{x}$ and $\Delta B_{y}$ respectively. The magnitude of the flow velocity is related to the magnitude of the magnetic field response, $\Delta B_{xy} = (\Delta B_{x}^{2} + \Delta B_{y}^{2})^{\frac{1}{2}}$. The orientation of the flow relative to the sensor is proportional to $\Delta B_{y/x} = \arctan{(\frac{\Delta B_{y}}{\Delta B_{x}})}$. For calibration, twenty data points were collected from each test. Other sensors can be calibrated using a similar approach if needed in the future. 

\begin{figure*}[t]
   \begin{center}
  \includegraphics[width=6in]{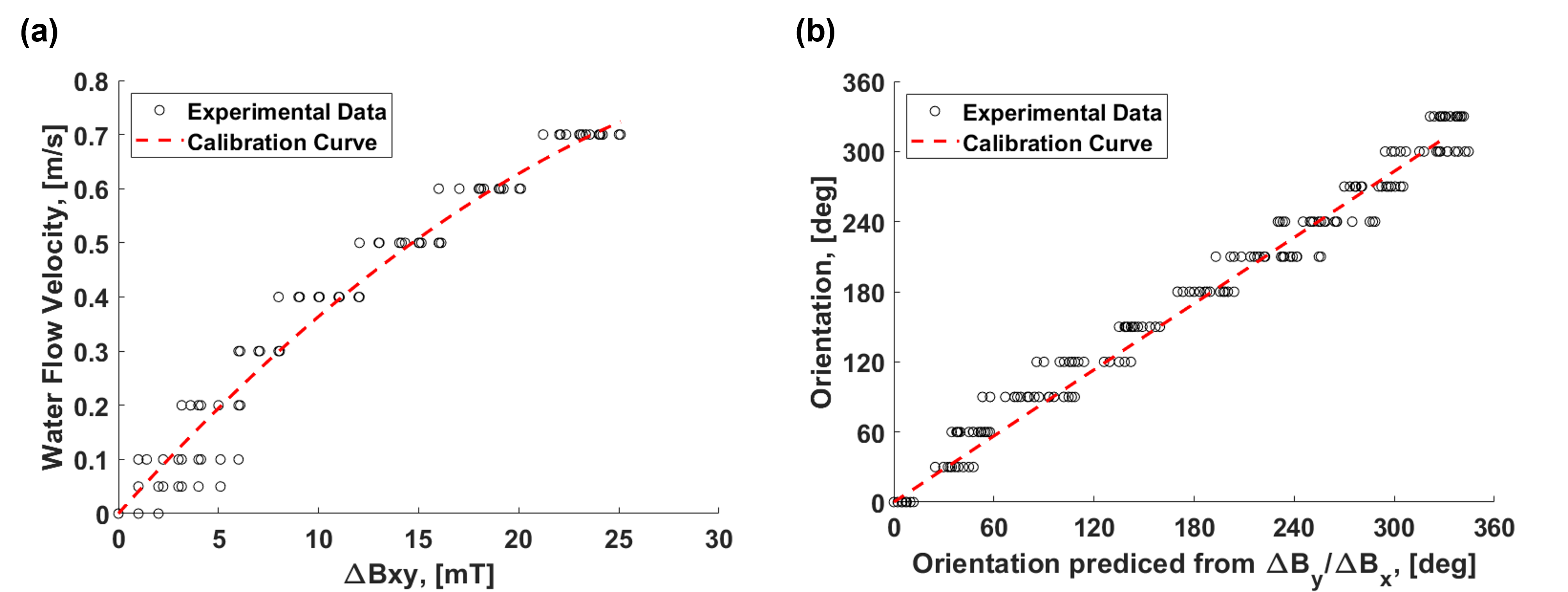}
  \caption{Calibration Results. (a) Calibration map of the magnetic field response to flow velocities form \SI{0}{\metre\per\second} to \SI{0.7}{\metre\per\second}. (b) Calibration map of the magnetic field response to flow orientations from \SI{0}{\degree} to \SI{360}{\degree}.}\label{vc}
  \end{center}
\end{figure*}

\section{Results}

\subsection{Varying whisker morphologies}

Both experiments and modeling were carried out to examine the magnetic field response of three whisker-inspired sensor shapes, aiming to understand their effectiveness in detecting water flow velocity and orientation. Fig. \ref{characterizationResults} shows the experimentally measured mean and standard deviation of the magnetic field response ($\Delta B_{x}$, $\Delta B_{y}$) for the three whisker shapes relative to water flow velocity and orientation. The x and y directions correspond to the Hall effect sensor's sensing directions. In Fig. \ref{characterizationResults}(a-c), the $\Delta B_x$ direction is in-line with the linear stage's movement, and flow in the transverse direction is measured by $\Delta B_y$ (Fig. \ref{characterizationSetup}(c)). This orientation corresponds to $\theta_{xy} = $~\SI{0}{\degree}. 

In terms of velocity characterization, all whisker shapes display an approximate quadratic relationship between the magnetic field response and flow velocity in the in-line direction. The transverse direction maintains an average near-zero response, consistent with the analytical model's predictions. This result signifies that the sensor's response to in-line motion can be separated from vibration in the transverse direction, enabling velocity estimation for all whisker shapes. For orientation characterization, rod whiskers show a nearly ideal sinusoidal magnetic field response in both x and y directions. The projected cross-sectional area ($A$) for cross and plate sensors depends on the flow direction, $\theta_{xy}$ (Eqn. \ref{referenceArea}). Cross whiskers exhibit an approximately linear relationship between the magnetic field response and orientation. Plate whiskers display minimal variation in magnetic field response along the y-axis, while the magnetic field for $\Delta B_x$ goes down to zero when the in-line direction shifts from the x-axis to the y-axis.

Vortex-induced vibrations (VIVs), a phenomenon where fluid flow causes vibrations in structures, is significant in the transverse direction measurements. Rod whiskers exhibit a bell-shaped VIV profile, peaking at \SI{0.3}{\metre\per\second} (Fig. \ref{characterizationResults}). In contrast, cross and plate whiskers show increased VIV magnitude in the high-velocity range, \SI{0.5}{\metre\per\second} - \SI{0.7}{\metre\per\second} (Fig. \ref{characterizationResults}(b-c)). Ultimately understanding VIV behavior is essential for optimizing whisker-based designs and mitigating vibration-related damage or fatigue in applications such as underwater sensing and bio-inspired robotics.


These experimental results demonstrate reasons for preferring one drag element design over another. For example, the plate whisker only picks up the component of flow in line with its x-axis and the cross whisker can go to higher velocities without significant VIV response. Once a researcher picks the shape best suited for their application, the model can help design the sensor for the velocity range needed. Experimental data from all nine whisker designs were separately compared to the model in Fig. \ref{Modeling}, yielding a prediction of the in-line $\Delta B_x$ value with a root mean square error (RMSE) of \SI{1.29}{\milli\tesla} for a given flow velocity. Given a measured $\Delta B_x$ value (as would be the case on an aquatic vehicle), the model's velocity prediction had an RMSE of \SI{0.034}{\metre\per\second}. The experimental results are highly repeatable and we suspect that most of this error is due to manufacture of the sensor's spring suspension element as discussed further in \cite{MRLwhisker01}.

\subsection{Varying immersion depth for rod whiskers}


Results of the characterization process for different immersion depths of rod whiskers are illustrated in Fig. \ref{depthcharacterization}(a-c). Notably, the in-line magnetic field response demonstrates an increase with the rise in flow velocity for all whisker diameters. Furthermore, the larger the diameter of the rod whisker, the higher the detected magnetic field response due to the increase in cross-sectional area which consequently leads to a higher drag force at the same flow velocity.

In addition, the magnetic response to flow velocity increases as the rod sensor's area enlarges through an increase in $D_{im}$. These results shed light on the affect of varying the immersion depth on the magnetic field response to changes in velocity without altering the whisker's design. Increasing $D_{im}$ both decreases the moment arm and increases the area of the whisker drag element (Eqns. \ref{MomentArm},\ref{dragEquation}). However, changing $D_{im}$ has a larger effect on the sensor's area versus the moment arm, leading to an increase in the magnetic field response (Fig. \ref{depthcharacterization}).

The VIV response exhibits a notable scaling with the immersion depth and is prominent in a fixed flow region for all whisker diameters. Figs. \ref{depthcharacterization}(d-f) show the magnitude and the velocity region where VIVs occur in the transverse direction. The experimental data reveals that the magnitude of VIV is strongly influenced by both the immersion depth and the rod diameter. Across all three rods, the VIV response is markedly pronounced for flow velocities from \SI{0.2}{\metre\per\second} to \SI{0.4}{\metre\per\second}.



Using the experimental data from the immersion depth tests, we evaluated the sensor model's ability to predict the sensor performance across three sensor performance metrics as described in Section II.A.4. Modeled and experimental results for the Hall effect sensor's sensitivity to changes in velocity, the sensed velocity in the middle of the sensor's range, and the maximum velocity a sensor can detect without damage to the sensor are plotted together in Fig. \ref{Design Decisions} for different whisker diameter and immersion depth variations. The model's predicted sensitivity was accurate to a root mean squared percentage error (RMSPE) of \SI{14.4}{\percent} across all of the immersion depth trials for the sensor's nominal design values (Fig. \ref{Design Decisions}a). The model's expected mid-point velocity estimate had an RMSPE of \SI{11.0}{\percent} across all trials (Fig. \ref{Design Decisions}b). The maximum tested velocity was \SI{0.7}{\meter\per\second} so modeling results for only two designs could be confirmed with experimental data (Fig. \ref{Design Decisions}c). The percentage error for these trials was \SI{14.1}{\percent} and \SI{16.2}{\percent} for immersion depths of \SI{30}{\milli\meter} and \SI{40}{\milli\meter} respectively. We use percentage errors over the RMSE to compare the model's predictive abilities over the three specifications whose values have different magnitudes.




We hypothesize that the nominal design values used to model the sensor are a large contributor to the model prediction error. 
When the spring width in the model is fit to the experimental results (as done in \cite{MRLwhisker01}), the percentage errors drop to \SI{1.38}{\percent} and \SI{3.96}{\percent} for the maximum velocity data points in Fig. \ref{Design Decisions}c respectively (these results were from the same sensor with the same spring at different immersion depths). Regardless, the trends remain consistent between the model and experimental results for all three specifications, and the model provides an approximate estimate which can lead to more informed sensor design choices. 

\subsection{Calibration Results}

Given manufacturing variations in sensor design, it makes sense to calibrate the sensor to improve its accuracy. A curve fixed at the origin was fit to the experimental data for flow velocity and orientation by using a least squares approach (Fig. \ref{vc}). The flow velocity and $\Delta B_{xy}$ were fit with a second order polynomial with an $R^2$ value of \num{0.969}. The orientation of water flow can be mapped to $\arctan(\Delta B_{x}/\Delta B_{y})$ in a linear relationship with an $R^2$ value of \num{0.981}.

The calibrated curve equations for flow velocity, $v$ (\si{\meter\per\second}), and orientation, $\theta_{xy}$ (\si{\degree}), for this sensor are listed below:

\begin{gather}
v = -0.0005(\Delta B_{xy})^2+0.0414(\Delta B_{xy}) \label{velocityCalibrationEqn}\\
\theta_{xy} = 0.944 \arctan(\Delta B_{y/x})
\label{orientationCalibrationEqn}
\end{gather}


    


\section{Application}

\begin{figure*}[t]
   \begin{center}
  \includegraphics[width=6in]{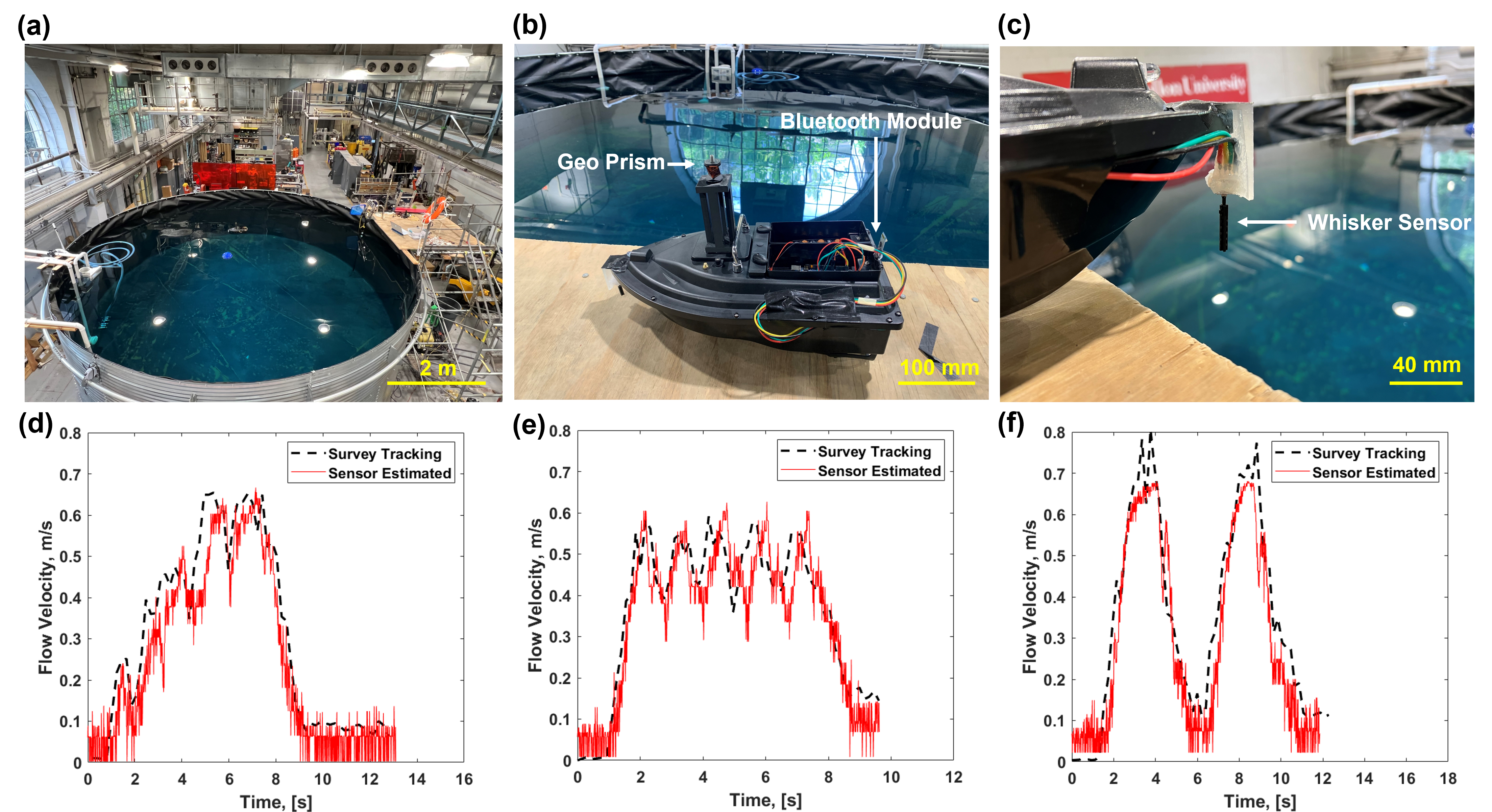}
  \caption{The test setup and results from on-board velocity estimation. (a) A water tank of \SI{8}{\metre} in diameter was used to test the whisker-inspired sensor on a RC boat. (b) The boat was tracked by a laser-prism survey system. The magnetic field response from whisker-inspired sensor was transmitted through Bluetooth. (c) The \SI{20}{\milli\metre}$\times$\SI{7.5}{\milli\metre} wide cross sensor was mounted on the head of the boat. (d) (e) (f) The boat performed three velocity profiles. The estimated velocity response from the whisker-inspired sensor was mapped with the ground truth velocity from the survey system. }\label{demo}
  \end{center}
\end{figure*}

\subsection{Test Setup}

In order to showcase the whisker-inspired sensor in a real-world aquatic environment, we integrated the chosen sensor on an RC boat (\SI{485}{\milli\metre} $\times$ \SI{264}{\milli\metre} $\times$ \SI{160}{\milli\metre}, D16C, Cresea Product) for velocity estimation. Given the prescribed boat velocity and tasks, a \SI{20}{\milli\metre} $\times$ \SI{7.5}{\milli\metre} cross-shape sensor was selected for this application. 

A joystick was used to remotely control the velocity and direction of the boat. Four wires from the sensor were taped along the edge of the boat to the holder at the back and connected with an Arduino microcontroller (Arduino Mega 2560 R3) through I2C communication. The sensor system was powered by a 9V battery pack. A Bluetooth module (HC-05, HiLetgo) was used to transmit the sensing data to a laptop. The boat with the sensor was tested in a \SI{8}{\metre} diameter, \SI{3}{\metre} high water tank at the Field Robotics Center (FRC) at Carnegie Mellon University. 

Figs. \ref{demo}(a-c) show the test setup. While the boat was operating in the tank, a survey system (Leica TS16) was used  to record the position and velocity of the boat as ground truth data. The survey system consists of a laser source and a geo prism (Leica GRZ101). During the test, the laser source actively tracked the geo prism and stored its positions in Cartesian coordinates with a sampling frequency of \SI{5}{\hertz}. 

\subsection{On-Board Test Results}

Figs. \ref{demo} (d-f) compare the velocity profiles estimated from the whisker-inspired sensor and the survey system. The motion of the boat was designed to provide a variety of velocity profiles within a range of \SI{0}{\metre\per\second} to \SI{0.8}{\metre\per\second}. The whisker-inspired sensor successfully captured all velocity variations with RMSE values of (d) \SI{0.08}{\metre\per\second}, (e) \SI{0.08}{\metre\per\second}, and (f) \SI{0.06}{\metre\per\second}. 
The velocity profile in Fig. \ref{demo}(d) was designed to have a peak velocity close to the limit of the sensor, and this profile was qualitatively captured by the whisker-inspired sensor. The velocity profile in Fig. \ref{demo}(e) was designed to have \num{5} periodic accelerations, and the whisker-inspired sensor successfully captured all of them. Notably, in Fig. \ref{demo}(f), the sensor output saturated at around \SI{0.7}{\meter\per\second}. This was deliberately designed by printing physical restrictions on the holder to prevent the sensor's spring suspension from plastic deformation. These findings are important as they indicate the sensor's ability to detect and measure complex dynamic events, emphasizing its potential applicability in various scenarios requiring accurate state estimation and motion tracking.

While using the calibration models (Eqns. \ref{velocityCalibrationEqn}, \ref{orientationCalibrationEqn}) for velocity and orientation estimation, we can also obtain the vortex-induced vibration signals from data not shown here. These signals are characterized by the time-dependent data from the direction perpendicular to the flow. For example, as the boat moves in the $B_{x}$ direction, VIV signals are obtained as time-dependent signals from channel $B_{y}$. This highlights the robustness and versatility of the approach in extracting valuable information from different channels.

\section{Conclusions}

This study presents the design of whisker-inspired sensors that can detect multi-directional water flow with a designed sensitivity. The sensitivity was tunable based on the designed whisker geometry, which we quantified using a model. We characterized how different variations in the whisker's shapes affected the magnetic field response in relation to flow velocity and orientation. Additionally, we investigated Vortex-Induced Vibrations (VIV) induced by different whisker structures at various flow velocities. Sensitivity was also tunable after manufacturing by modifying the immersion depth which is possible for surface aquatic vehicles. Our model also demonstrated the ability to design a sensor for a specific range of flow velocities. Once the sensor was manufactured, we improved the accuracy of the velocity prediction from the magnetic field by calibrating the model to the sensor's final parameters. 

To test the capabilities of the whisker-inspired sensor, we implemented it on a commercially available RC boat and demonstrated its ability to estimate velocity in a static water environment. We then compared our estimated velocity with ground truth data captured by a survey system. The onboard sensor configuration and wireless Bluetooth data transmission make this whisker-inspired sensor system versatile, offering potential for use in various environments, including outdoor fieldwork, remote locations, or situations where WiFi access is restricted or unavailable.



\section*{Acknowledgment}

The authors would like to thank Regan Kubicek from Carnegie Mellon University for early help with experimental setup and advice on the fabrication process, along with Suhan Kim from MIT for advice on data acquisition and inspiration. The authors also would like to thank Prof. Mitra Hartmann and Kevin Kleczka from Northwestern University for inspiration and feedback on sensor characterization. Finally, the authors would like to thank Prof. Michael Kaess and Warren Whittaker from Carnegie Mellon University for their generous help and training for the use of the water tank. This work was partially supported by  MURI award number FA9550-19-1-0386.

\bibliographystyle{IEEEtran}

 
\vspace{11pt}

\begin{IEEEbiography}[{\includegraphics[width=1in,height=1.25in,clip,keepaspectratio]{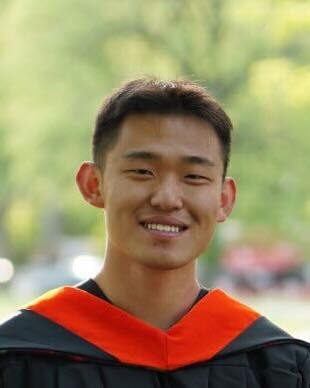}}]{Tuo Wang}
(IEEE, ASME Student Member) received the B.S. degree in Mechanical Engineering from the University of Pittsburgh, Pittsburgh, PA, USA, in 2021, and the M.S. degree in Mechanical Engineering from Carnegie Mellon University, Pittsburgh, PA, USA, in 2023. He currently works as a Post-Baccalaureate Research Fellow with the SeNSE lab, Northwestern University, Evanston, IL, USA,  on bio-inspired flow sensors. His research interests include bio-inspired robotics, wearable devices, and computer-aided engineering.
\end{IEEEbiography}

\begin{IEEEbiography}[{\includegraphics[width=1in,height=1.25in,clip,keepaspectratio]{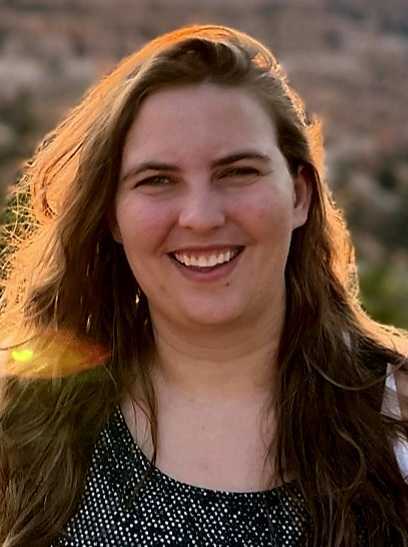}}]{Teresa A. Kent}
(IEEE Student Member) received a B.S. degree in Mechanical Engineering from the University of Maryland College Park, College Park, MD in 2017, and an M.S. in Mechanical Engineering from Carnegie Mellon, Pittsburgh, PA in 2019. She is currently a Ph.D. candidate at Carnegie Mellon University, Pittsburgh PA in the Robotics Institute. Her research interests include sensor design, whisker sensing, tactile sensing, applications for computer vision, and soft robotics.
\end{IEEEbiography}

\begin{IEEEbiography}[{\includegraphics[width=1in,height=1.25in,clip,keepaspectratio]{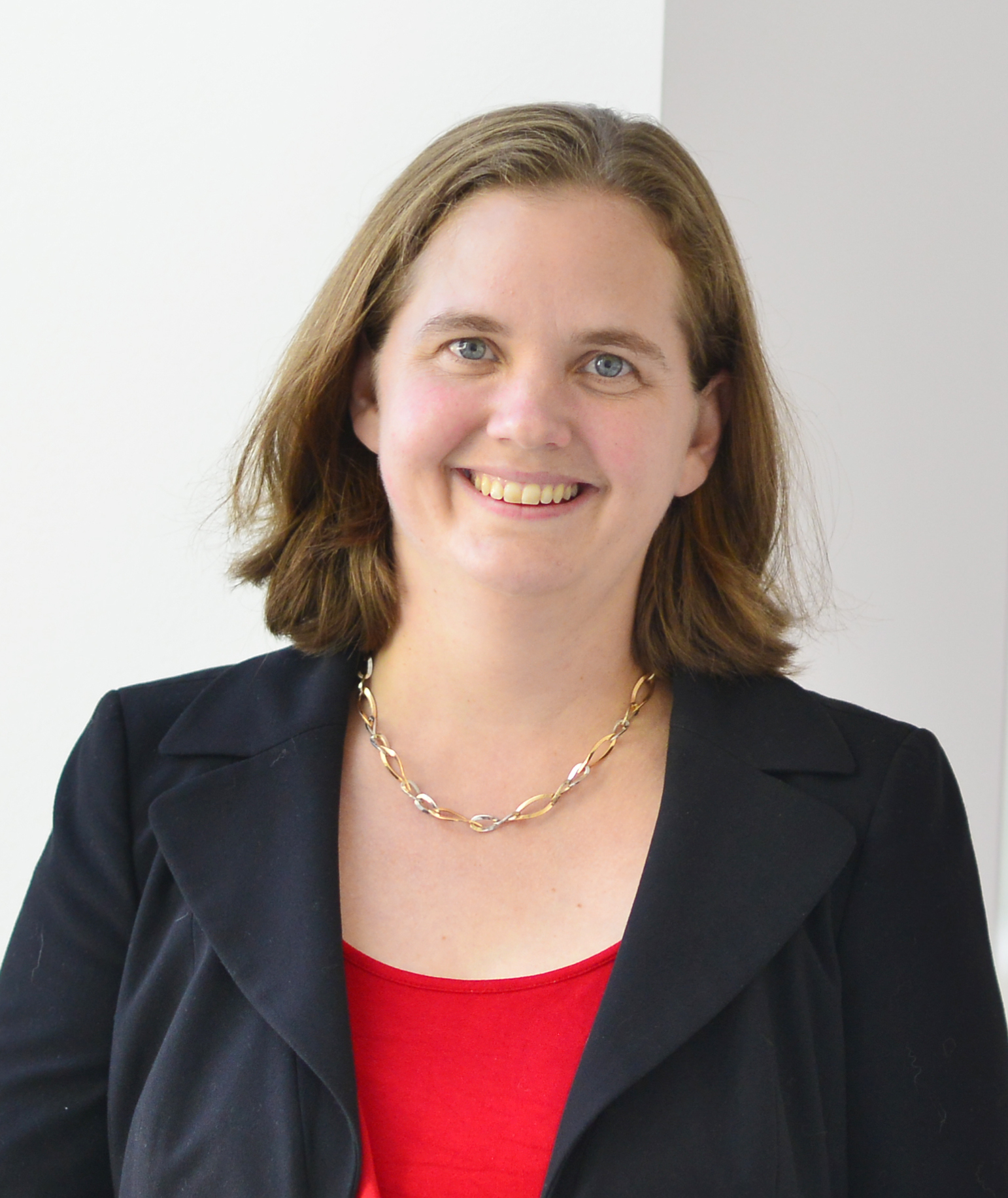}}]{Sarah Bergbreiter}
(ASME Fellow, IEEE Member) joined the Department of Mechanical Engineering at Carnegie Mellon University as a Professor in the fall of 2018 after spending ten years at the University of Maryland, College Park. She started her academic career with a B.S.E. degree in electrical engineering from Princeton University in 1999. After a short introduction to the challenges of sensor networks at a small startup company, she received the M.S. and Ph.D. degrees from the University of California, Berkeley in 2004 and 2007 with a focus on microrobotics. Prof. Bergbreiter received the DARPA Young Faculty Award in 2008, the NSF CAREER Award in 2011, and the Presidential Early Career Award for Scientists and Engineers (PECASE) in 2013 for her research on engineering robotic systems down to millimeter size scales. She has received several Best Paper awards at conferences like ICRA, IROS, and Hilton Head Workshop.
\end{IEEEbiography}



\vfill

\end{document}